\documentclass[12pt]{article}

\usepackage{scicite}

\usepackage{times}

\usepackage{amsmath, bm}
\usepackage{amssymb}
\usepackage{bbm}
\usepackage{lineno}
\usepackage{url}

\usepackage{graphicx}
\usepackage[section]{placeins}

\newenvironment{sciabstract}{%
\begin{quote} \bf}
{\end{quote}}

\title{Tactile-Reactive Roller Grasper}

\author
{Shenli Yuan$^{1\ast\dagger}$, Shaoxiong Wang$^{2\dagger}$, Radhen Patel$^{2}$, Megha Tippur$^{2}$, \\Connor Yako$^{1}$, Edward Adelson$^{2\ddagger}$, Kenneth Salisbury$^{1\ddagger}$\\
\\
\normalsize{$^{1}$Stanford University, Stanford, CA 94305, USA}\\
\normalsize{$^{2}$Massachusetts Institute of Technology, Cambridge, MA 02139, USA}\\
\\
\normalsize{$^\ast$To whom correspondence should be addressed; E-mail:  shenliy@ccrma.stanford.edu.}\\
\normalsize{$\dagger$ These authors contributed equally to this work}\\
\normalsize{$\ddagger$ These authors contributed equally to this work}
}

\date{}

\begin{document} 


\baselineskip24pt


\maketitle


\begin{sciabstract}
  
  Manipulation of objects within a robot's hand is one of the most important challenges in achieving robot dexterity. The ``Roller Graspers" refers to a family of non-anthropomorphic hands utilizing motorized, rolling fingertips to achieve in-hand manipulation. These graspers manipulate grasped objects by commanding the rollers to exert forces that propel the object in the desired motion directions. In this paper, we explore the possibility of robot in-hand manipulation through tactile-guided rolling. We do so by developing the Tactile-Reactive Roller Grasper (TRRG), which incorporates camera-based tactile sensing with compliant, steerable cylindrical fingertips, with accompanying sensor information processing and control strategies. We demonstrated that the combination of tactile feedback and the actively rolling surfaces enables a variety of robust in-hand manipulation applications. 
  In addition, we also demonstrated object reconstruction techniques using tactile-guided rolling. A controlled experiment was conducted to provide insights on the benefits of tactile-reactive rollers for manipulation. We considered two manipulation cases: when the fingers are manipulating purely through rolling and when they are periodically breaking and reestablishing contact as in regrasping. We found that tactile-guided rolling can improve the manipulation robustness by allowing the grasper to perform necessary fine grip adjustments in both manipulation cases, indicating that hybrid rolling fingertip and finger-gaiting designs may be a promising research direction.

\end{sciabstract}




\section*{Introduction}



Similar to how the dexterity of human hands allows us to accomplish a variety of everyday tasks, the in-hand manipulation capabilities of robots are necessary to accomplish a wide range of complex tasks in different environments. Out of all the manipulation tasks a human hand could perform, in-hand manipulation requires the most dexterity~\cite{bullock2011classifying}. To achieve such dexterity, it is only natural that researchers sought to create and use devices of similar function and construction to the human hand. Recent research has shown this to be a promising direction, with a simulated 20 degree-of-freedom (DoF) Shadow Hand~\cite{company2003design} able to manipulate in-hand over 2000 objects in a variety of palm poses~\cite{chen2022system}. However, while having high dexterity is desirable for robot hands, performing in-hand manipulation with robot hands or fingers based on their human counterparts may not be optimal in all situations. For example, legged quadrupeds have demonstrated great ability to navigate wide ranges of terrain in recent years, however, a quadruped's efficiency (in terms of task completion speed) on smoother surfaces can augmented by the inclusion of wheeled feet, evidenced by the wheeled ANYmal~\cite{bjelonic2019keep} and its predecessor from company Swiss-Mile. 
To that end, it may be prudent to pursue similar embodiments but for efficiently realizing certain dexterous in-hand manipulation tasks.



This mindset led to the development of the Roller Graspers \cite{rgv1,rgv2}, which introduced an entirely new way to manipulate objects within hand. They have demonstrated success in various in-hand manipulation applications. These include manipulation tasks of various objects such as a piece of paper, a die, and a 3D printed cube using continuous rolling, completed through open-loop control and through closed-loop control with the help of QR tags. 
However, one of the major limitations of the previous works is the lack of local contact information. Tactile sensors are a crucial component for successful robot in-hand manipulation as has been shown with various linkage-based robot hands~\cite{yousef2011tactile,fazeli2019see,she2019cable}. 
This paper details our investigation into the utility of robot in-hand manipulation through closed-loop tactile-guided rolling. We developed the Tactile-Reactive Roller Grasper (TRRG) shown in Fig.~1(A), a robotic grasper that utilizes camera-based tactile sensing and steerable rolling fingertips to perform a variety of manipulation tasks. The developed sensor information processing allows us to extract necessary contact information that can be used in real-time feedback control. The advantages of combining the Roller Grasper and the GelSight sensor are two-fold. First, incorporating tactile sensing greatly improves the in-hand manipulation capabilities of the Roller Grasper, by enabling the grasper to detect and exploit local contact information between the rollers and the grasped object to perform stable, and robust in-hand manipulation. Second, the steerable rollers enable the tactile sensor to easily scan potentially large and complex surfaces, leading to efficient and accurate 3D reconstructions. Our robotic system includes the TRRG, a Universal Robots UR5e robot arm, and a computer, as shown in Fig.~1(B). Aside from the physical design of the hand, we also addressed the sensor calibration and processing techniques related to the continuous rolling nature of the sensor. We formulated our control algorithm based on differential geometry and the availability of tactile information. 
The TRRG completed various tasks with the help of tactile-guided rolling such as tracing a flexible cable bidirectionally, imparting screw motions to a clear acrylic sphere, performing robust object reorientations, and picking up a single card from a deck of cards. In addition to the demonstrations, we performed an experiment to compare manipulations through rolling and regrasping, with and without tactile feedback. We found that while rolling appears to be more stable in the open-loop case, performing fine grip adjustment with tactile-guided rolling enabled stable manipulation for both the regrasping and pure rolling manipulation cases.





\subsection*{Review: Robot Hand for In-Hand Manipulation}

For the past century, there have been a number of robotic hands designed with in-hand object manipulation capability~\cite{piazza2019century}. Most of these hands are linkage-based, including anthropomorphic hands~\cite{company2003design,diftler2011robonaut,grebenstein2011dlr} and other fully-actuated hands~\cite{salisbury1982kinematics,jacobsen1984utah}, many of which have shown promising in-hand manipulation capability. 
Alternatively, in-hand manipulation can be achieved without intentionally switching contact locations, but this severely limits the object's range of motion. There have also been works that use underactuated linkage-based hands~\cite{ma2014underactuated,bircher2019energy} to achieve in-hand manipulation. 
Another approach towards in-hand manipulation is to use non-anthropomorphic hands~\cite{ma2016spherical,mccann2017design}. One particular type of non-anthropomorphic hands use active surfaces that allow the hand to move a grasped object without the need to lift its fingers. Earlier works on this type of robotic hand have fixed conveyor directions~\cite{datseris1985principles,govindan2019design,tincani2012velvet,ma2016hand,kakogawa2016underactuated}. \cite{rgv1} and \cite{rgv2} further developed this concept and incorporated steerable rollers for more dexterous in-hand manipulation. However, as mentioned before, the challenge of incorporating traditional tactile sensors into active surfaces made us choose a vision-based tactile sensor for this application.

\subsection*{Review: Vision-based Tactile Sensing}

Vision-based tactile sensors~\cite{ward2018tactip,alspach2019soft,sferrazza2020learning,yamaguchi2016combining,lambeta2020digit,padmanabha2020omnitact,romero2020soft} are a type of tactile sensor which converts contact signals into images. It has become increasingly popular in recent years because it provides high-resolution, force-sensitive data that are low-cost and flexible to modifications. Vision-based tactile sensors usually consist of a piece of elastomer, a camera, and a lighting system. When externally in contact with an object, the sensor captures the deformation of the elastomer by a camera and infers characteristics, such as the shape of the contact, and the shear and torsional forces.

While tactile sensors based on resistance, capacitance, and piezoelectricity~\cite{dahiya2009tactile,yousef2011tactile,cutkosky2016force} can be great options for regular linkage-based robotic hands, they are not suitable to be integrated with continuously rolling mechanisms. Because their design involves deploying wires/cables from the electronics to the sensing area, the wires/cables would inevitably get tangled due to the continuous rotation. Vision-based tactile sensors, in comparison, provide great advantages by allowing the tactile signal to be transferred through light, eliminating the mechanical coupling between the sensing area and the electronics making it an ideal choice for continuously rolling mechanisms.

There have been previous works~\cite{shimonomura2019tactile,cao2021touchroller} that integrate vision-based tactile sensors into passive rollers for inspection tasks. It was demonstrated that rolling action greatly improves the efficiency of inspection especially when scanning large areas. However, since the rollers in these works are passive, they rely on motions of the robotic arm to rotate, and cannot impart motions to an external object; thus, these systems do not have in-hand manipulation capabilities.

In this work, we integrate a category of vision-based tactile sensors - known as GelSight sensors~\cite{yuan2017gelsight} - into actively-driven rollers. This enables more robust in-hand manipulation through closed-loop rolling contact, and facilitates efficient inspection of the geometric properties of a grasped object during manipulation. In addition to the shear forces and 2D contact geometry, GelSight sensors can also provide high-resolution 3D contact geometry by applying photometric stereo~\cite{yuan2017gelsight}. The 3D information can be further processed and be used for normal forces estimation, pose estimation, and surface reconstruction. Contrary to~\cite{yuan2017gelsight}, we have redesigned the sensor's form-factor to fit into the compact design of the actively driven rollers and designed corresponding information processing pipeline, all while preserving the high-resolution 3D contact geometry. 

\section*{Results}


\subsection*{Sensor and Hand Design}

The TRRG is a two-fingered grasper with each finger consisting of three actuated DoF. The design of the grasper is shown in Fig.~2. The base DoF is driven by a Robotis Dynamixel XM430-W350 actuator through a four-bar parallelogram linkage. The mechanism enables up to $160$ mm opening between the rollers. The finger can exert normal forces between $0$ N and $68.3$ N, with the maximum force exerted when the input link of the four-bar linkage is vertical (fully closed grip). A micro DC motor embedded in the L-shaped hub controls the second DoF and is capable of pivoting the roller head between $\pm 90^{\circ}$ through a five-bar parallelogram mechanism seen in Fig.~2(A). The mechanism was improved from that in our previous work~\cite{rgv1} to allow for a greater range of motion. Another micro DC motor, shown in Fig.~2(B), is embedded at the back of the roller head to drive the roller through spur gears. Unlike the previous generations of the Roller Grasper, the roller motor is located external to the roller to make space for the tactile sensing components. The physical dimensions of the TRRG are presented in Table. 1, and a link to the complete CAD assembly is provided in the supplementary material. 

One of the most important considerations for vision-based tactile sensors is designing for clear optical paths. 
As a result, the mechanical structure of the roller consists of a clear acrylic tube glued with two clear acrylic rings on both ends, allowing unobstructed light passage between the light source, sensing area, and camera  (Fig.~2(C) and Fig.~2(D)). A 3D-printed gear is attached to one of the acrylic rings. Eight dowel pins arranged around the perimeter are inserted into both the gear and the acrylic ring to provide sufficient torque transmission. The clear gel (elastomer) is molded directly over the acrylic tube for ample optical transparency. 

The finger and the elastomer are designed to be as compact as possible to be better suited for manipulation of daily objects. The lower-limit of the size is constrained by the components located inside the roller with the fundamental limit being the focal length of the Raspberry Pi camera. As shown in Fig.~2(D), the camera is located at the bottom of the stator and angled at $20^{\circ}$ from its horizontal mounting surface. It streams images from the sensing area through a mirror oriented $20^{\circ}$ from the rolling axis.
The purpose of the mirror is to increase the distance between the sensing surface and the camera, which allows us to use a smaller acrylic tube when the camera focal length is fixed. The thickness of the elastomer is determined based on the following constraints: (1) it needs to be close the width of the LED array to allow unobstructed light passage; (2) it needs to be thin enough so that the directional light can illuminate the necessary surface area; (3) it needs to be thick enough to ensure the grasped object is in contact with the compliant roller rather than the rigid housing.
With this design, the resulting tactile sensor has a sensing area that forms a $90^{\circ}$ angle to the axis of the roller, which is equivalent to a roughly $33$ mm $\times$ $56$ mm area when stretched to a plane.

\subsection*{In-Hand Manipulation with Tactile Sensing}


We developed a series of demonstrations in order to highlight different aspects of the TRRG system. With prior knowledge that manipulation through rolling can provide good dexterity, these demonstrations will focus on how the integration of tactile sensing can provide the system with additional capabilities and robustness. 

In terms of kinematics, the two-finger design with six total actuated DoF for the TRRG is a significant simplification compared to the previous three-finger Roller Grasper V1/V2. However, the inclusion of tactile sensing allowed the TRRG to stably manipulate objects without the extra redundancies necessary for grasp stability in previous generations.
Even with the vastly reduced DoF, the TRRG is still capable of translating or rotating the grasped object in each of the $X_O$, $Y_O$ and $Z_O$ directions, as defined in Fig.~3(A). By utilizing the combinations of the various manipulation primitives presented in Fig.~3(B-F), grasped objects can be manipulated between a wide variety of initial and target poses.

The tactile sensor provides both depth and shear information for an object in contact with the rollers, and the raw sensor data can be further processed to extract higher-level information suited for in-hand manipulation. The contact location can be used to close the low-level control loop for determining each finger's rolling speed and direction (Fig.~4(A-C)). The depth information along with the shear information can also be used to inform the high-level control loop in order for the hand to perform more comprehensive tasks robustly (Fig.~4(D)(E)).

The tactile sensor also helps mitigate certain hardware limitations. 
Specifically, since the TRRG acts as a parallel jaw grasper it needs to apply sufficient normal forces to stably grasp objects. To ensure that the rollers are still able to roll even in the presence of large grasp forces, we calculated the necessary output torque to the roller based on a highly conservative rolling resistance coefficient of $0.4$ and the maximum normal force able to be applied by the roller to the object, determined to be $68.3$ N. Note that a rolling resistance coefficient of $0.2 - 0.4$ is used to describe a car tire attempting to roll on loose sand. With the current design, the required gear ratio for the roller is $331$, making this DoF mechanically non-backdrivable.
However, the tactile sensor enables force control along the shear direction.
The rollers can either not react to the shear force (as shown in Fig.~4(F)(i)), taking advantage of the friction of the transmission for secure grasping, or actively adjust the roller speed based on the shear information (as shown in Fig.~4(F)(ii, iii)), allowing for compliant manipulation despite having a non-backdrivable transmission. Further details on the individual demonstrations as well as the formulation of the controller are presented in the Materials and Methods section. 




\subsection*{Efficient object/image reconstruction using steerable rollers}

The rolling action combined with tactile sensing results in efficient surface inspection or reconstruction, compared with previous methods where researchers needed to apply a sequence of discrete touches when inspecting surface roughness after sanding\cite{amini2020uncertainty}, detecting defects on objects \cite{fang2022fabric,jiang2021vision}, and reconstructing 3D shapes\cite{bjorkman2013enhancing,wang20183d,smith20203d}.


We demonstrated this ability through the reconstruction of surface geometries for both a credit card (2D) and a transparent cup (3D). In the 2D demonstration, the grasper carefully guided the credit card in between its rollers while scanning the surface textures of the card, as shown in Fig.~5(A-C).
The tactile images were then stacked in the time sequence to recover the credit card numbers.
We also show the raw stacked images in comparison with the processed images to better visualize the effects of filters and the interpolation required to fill the areas occluded by the black markers.

The 3D demonstration reconstructed a transparent cup with an embossed logo. Unlike ~\cite{cao2021touchroller}, the greatly reduced form-factor of our rollers allow us to reconstruct much more complex geometries. The cup was mounted on a turntable to enable pure rotation around $Z_O$. 
The rollers held onto the opposing sides of the cup near its opening, and traversed from the lip to its base; the rollers then tilted a small angle ($5^{\circ}$) and rolled upwards until they reached the opening of the cup at adjacent positions to where they started. The slightly tilted angles of the rollers resulted in a screw motion of the cup relative to the grasper: every time when the rollers rolled down and up, the cup was rotated by a small angle. The tilted angle needs to be small enough to ensure that the entirety of the surface is scanned with each vertical pass. 
This sequence repeats until the scanning of the entire surface is completed. To better align the multiple scanned images, we applied cross-correlation between them to match their correspondence. The resulting images are presented in point clouds, as shown in Fig.~5(D-F). We specifically selected a transparent cup to show the capabilities of this technique, which can be challenging to accomplish when using color or depth cameras.

\subsection*{Tactile-guided rolling for regrasping and rolling manipulation}

We designed an experiment to investigate manipulating an object through rolling and through regrasping, with and without the use of tactile sensing as feedback. Regrasping was investigated because the TRRG only has two fingers, and is not suitable to perform finger-gaiting. While regrasping is not as common as finger-gaiting, the underlying principle is very similar to that of finger-gaiting; both involve a periodic lift and replacement of the fingers at a different locations on the object, however, regrasping requires each finger to be lifted simultaneously from the object; in these situations the object is typically on a support surface. In finger gaiting, only a subset of the fingers will be lifted at a given time. Comparing regrasping and manipulation through rolling with respect to the same task allows us to gain an understanding into the interplay between continuous rolling, breaking and reestablishing contact, and tactile sensing. 

The experiment is designed to rotate a rectangular prism around its vertical axis using the TRRG and the wrist of a UR-5e. The object is placed on a support surface throughout all experiments and is initially positioned between the two rollers. We used the tactile sensor to record the contact location between the object and the roller in all of the experimental conditions listed below in order to observe the evolution of the contact location during the manipulation. Based on the recorded tactile data and the supplemental video (Movie S2), we can also calculate the total object rotation in each experiment before the object was lost from the grasp. 

There are four experimental conditions are as follows:
\begin{itemize}
    \item \textbf{Regrasp open-loop}: the grasper would periodically grip the object, rotate the object by rotating its wrist for 0.3 rad, then release the grip and reset to its initial position. The rollers are not operating in this condition.
    \item \textbf{Regrasp closed-loop}: in addition to all the actions taken above, while the wrist is rotating one of the rollers may rotate based on the sensed contact location to fine-tune the grip pose. 
    \item \textbf{Roll open-loop}: the grasper rotates the object by rolling without feedback from the tactile sensor.
    \item \textbf{Roll closed-loop}: the grasper rotates the object by rolling while simultaneously adjusting the rolling speed based on the contact location extracted from the tactile sensor.
\end{itemize}

 In each condition, the horizontal contact location in image pixels, $v_s$, is set to a desired value $v_s=0$, which represents the contact location being centered on the roller surface from the perspective of the camera. The results of the experiment can be seen in Fig.~6(A), which shows the evolution of the contact location of the object on the roller during manipulation
 Video recordings of the experiment are provided in the Supplementary Materials. Without tactile feedback the grasper eventually loses control of the object during manipulation.
The results from the regrasp conditions also show that each time when the grasper reestablishes contact on the object, the contact location has shifted from its previous position. This type of uncertainty appears to make the regrasping method less stable compared to continuous rolling when tactile feedback is not used, as the regrasp method lost control of the object after less than 0.25 revolutions, while the rolling method rotated the object for 2.5 revolutions before losing grasp. 



Regarding the closed-loop experiments, the addition of tactile sensing allowed both the regrasping and the continuous rolling cases to successfully manipulate the object stably, without losing it from the grasp. For the pure rolling case the tactile information could be used to continuously monitor and adjust the grasp, shown by the smooth orange curve in the bottom plot of Fig.~6(A). However, even in the regrasp case where the tactile information was only available intermittently, the unique combination of rolling and tactile sensing allowed the grasper to perform successful fine grip adjustments to keep the grasp highly stable throughout the manipulation. Therefore, regardless of whether an object is manipulated by pure rolling or a combination of rolling and regrasping, it appears as though tactile-guided rolling for fine adjustment may always be beneficial, and possibly necessary to perform safe and successful in-hand manipulation.

\section*{Discussion}
\subsection*{Contributions}


This work investigates how in-hand manipulation can be augmented by use of the tactile sensing in conjunction with rolling contacts. To this end, the paper presents the design of the Tactile-Reactive Roller Grasper that has steerable active rollers at fingertips integrated with high-resolution tactile sensors. We developed algorithms to process tactile signals in order to provide real-time feedback for in-hand manipulation and object reconstruction. A generic low-level controller was formulated based on differential geometry to use the tactile information to perform in-hand object manipulation, which is suitable for all the different form-factors of this class of graspers that have been previously proposed. We also demonstrated more comprehensive tasks by closing the higher-level control loop using only the tactile information. 
Through the mechanical and algorithmic design of the TRRG, it demonstrated the ability to perform robust in-hand manipulation for various objects through tactile-guided rolling contact, even with unknown dynamics or external disturbances. We also demonstrated its unique capability to perform efficient surface inspection and reconstruction of surface geometries during manipulation. This was only made possible with the combination of actively driven rolling contact and high-resolution tactile information. 
In addition to the various demonstrations, we conducted an experiment to investigate manipulation through rolling vs. regrasping, and the effect of using tactile sensing as feedback. We found that when tactile feedback was used to perform fine grip adjustments, both manipulation through rolling and manipulation through regrasping were markedly more stable. This suggests that even devices typically constrained to pure finger-gaiting could benefit from the inclusion of roller-based fingertips to finely adjust the grasp without the need for lifting and re-placing the finger on the object.

In summary, we presented the abilities of the TRRG for in-hand manipulation and object reconstruction as well as its potential to complete complex perception and manipulation tasks in various real-world robotic settings. We hope this work would push the boundaries in both robotic manipulation and tactile sensing, and inform design decisions for future work in tactile-guided rolling manipulation.

\subsection*{Limitations and future works}

While we have demonstrated the incredible abilities of the TRRG, there are different aspects of this work that can be further explored.

In terms of the design of the grasper, although the rollers on the TRRG have a convex curvature, a spherically shaped roller allows for more consistent contact behavior. As noted in~\cite{rgv1}, when rollers with large radii of curvature are contacting objects with large radii of curvature the contact point can shift unpredictably in the face of minor misalignment.

Another limitation of the current design is that the size of the sensing area is restricted by the $90^{\circ}$ camera field of view as well as the mirror size and shape, which could be increased by using a camera with large FOV (e.g., a fish-eye camera) and a convex mirror. 

Our demonstrations presented feedback control methods using only tactile information, however, inclusion of additional sensing modalities such as force sensing and visual feedback could provide both global object information as well as local contact information. 

The TRRG's control pipeline could be augmented through integrating its object geometry reconstruction and in-hand manipulation abilities: while the TRRG can manipulate objects with unknown geometry and dynamics, the geometry of the object reconstructed during manipulation can further be used to improve the manipulation results.


Lastly, while maintaining contact during manipulation demonstrates many benefits, there are certain situations when lifting the finger is desired during manipulation. For example, when the object has sharp protrusions or deep concave features, manipulation through rolling may no longer be an optimal solution. A more comprehensive controller that accounts for such extreme cases would increase the diversity of the objects being manipulated and further improve the robustness of manipulation.

\section*{Materials and Methods}

\subsection*{Sensor fabrication}

\textbf{Lighting}
The lighting system was designed to have accurate 3D reconstruction based on photometric stereo, while fitting into the compact form-factor of the roller. To satisfy the requirements, we modified the design of the lighting system from the GelSight Wedge sensor~\cite{wang2021gelsight} to be suitable to illuminate the curved roller surface.

As shown in Fig.~2, the clear acrylic tube located at the center of the roller provides mechanical support for the rotor while allowing light to shine through. A camera is mounted at the bottom of the stator, and captures the sensing area through a mirror. Two LED bars (one blue and one red, respectively) are located near either vertical edge of the mirror to provide directional light from two different directions toward the sensing area. A green LED ring attached below the roller shines light through the bottom acrylic plate to provide the third color component essential for 3D reconstruction. Clear UV resin connects the clear acrylic plates to the clear acrylic tube, keeping the interface optically clear for the light from the LED ring.


\textbf{Camera}
We used a Raspberry Pi camera with a $120^{\circ}$ FOV, allowing us to obtain a relatively large sensing area while fitting the camera inside the tight interior of the roller. The camera was customized with a $200$ mm long flex cable, so the bulky connector can be located outside the roller. We streamed the video at $30$ Hz through \textit{mjpg\_streamer} to the Raspberry Pi, with a $640\times480$ resolution. The images were then transmitted from the Raspberry Pi to a PC for further processing. We were able to achieve a $30$ Hz update rate for the processed sensor signals; the update rate is highly dependent on the number of iterations used for the marker tracking optimization. 

\textbf{Elastomer} 
We designed and fabricated the seamless elastomer to obtain continuous tactile signals during rolling. 
In comparison, another fabrication technique is to cast a piece of flat elastomer to be wrapped around the rotor core\cite{shimonomura2019tactile, cao2021touchroller}, which would be less durable and result in discontinuous sensing signals at the seam. 

Fig.~7(A) shows the sequence of the elastomer fabrication. We first 3D printed the positive mold and smoothed the curved surface with a layer of coating. A stretchable negative mold was then cast using the translucent silicone. Next, the clear silicone on the roller was cast together with the acrylic tube. We applied a layer of primer on the outside of the acrylic tube before casting to increase the acrylic-to-elastomer bonding. Finally, we sprayed a layer of opaque gray (Lambertian) silicone inks onto the surface of the roller. 



\textbf{Markers}
To provide information of shear and torsional forces, we added multiple arrays of markers on the surface of the roller. The markers were created on a lasercutter on with a rotary attachment, which etched away the gray coating at the pre-defined marker locations. The rollers were then applied with a layer of black silicone ink, resulting in black makers with a gray background in the camera view.

The density of the markers is determined based on the desired feedback rate as marker tracking is the most computationally intensive part in our sensor signal processing pipeline. As mentioned previously, the number of markers used in the TRRG allows the tactile information to be updated at the rate of $30$ Hz. For other task specific designs, a higher density will lead to higher resolution measurements of the force fields at the cost of higher computational requirements to track individual markers.

\subsection*{Tactile Signal Processing}
This section discusses the signal processing techniques for the raw tactile signal. It also addresses the challenges created by the continuous rolling and convex sensing surface, and the corresponding solutions that we proposed.

\textbf{Encoding}
In order to achieve 3D reconstruction and marker tracking, the signal processing algorithms require each image to be compared with a reference image taken in the absence of contact~\cite{yuan2017gelsight}. Unlike GelSight sensors with the conventional form-factor, our sensing area expands the entire perimeter of the roller and thus multiple reference images in correspondence to different roller positions need to be taken in order to properly process the sensor signal. This requires the algorithm to find the correct reference image. However, due to backlash in the transmission and latency between the actuator and camera, the roller motor encoder cannot be used to correspond a given image to its designated reference. Therefore, we attached an encoder inside the camera FOV, as shown in Fig.~7(C), in order to match a given image with its reference for real-time tracking and minimal sensor noise. The encoder designed with this method can achieve pixel-level precision.

During the calibration process, the roller slowly rotates at a constant speed, allowing the camera to record reference images along with the encoder images in order to construct a lookup table for each frame. During manipulation, we extract the encoder portion of the image and find the L2 distance between the current encoder image and references from the lookup table to determine the corresponding reference image. Finding the correct reference image is a crucial early step toward the successful processing of tactile signals. 

\textbf{Surface Projection}
Camera matrices are used to calculate the correspondence between the points on the sensor surface in 3D and the 2D camera image pixels.
The transformation~\cite{szeliski2010computer} can be represented as:
\begin{equation}
\lambda
\begin{bmatrix}
           v_s \\
           u_s \\
           1
         \end{bmatrix} = K [R|t] \begin{bmatrix}
           X_5 \\
           Y_5 \\
           Z_5
         \end{bmatrix}
\end{equation}
where $\bm{u}_s = (u_s, v_s)^T$ represents the image coordinates of the sensor input; $\lambda$ is a scale factor; $K$ is the camera intrinsic matrix; $[R|t]$ is the camera extrinsic matrix, with rotation $R$ and translation $t$; $(X_5, Y_5, Z_5)^T$ represents the 3D coordinates in the sensor frame, shown as Frame $\mathcal{A}_5$/$\mathcal{B}_5$ in Fig.~3(A). Note that unlike $\mathcal{A}_6$/$\mathcal{B}_6$, which are fixed to the rollers, $\mathcal{A}_5$/$\mathcal{B}_5$ are fixed to the cameras or the stator inside the roller, which are not rotating with the rollers.

The camera was calibrated using a 7x8 checkerboard.
The camera, along with the mirror, was first mounted to the 3D printed housing and calibrated before the stator was assembled with the rest of the roller head.
During camera calibration, multiple sensor images were collected with different checkerboard poses, which were later used for providing the camera intrinsic matrix $K$.
The extrinsic matrix $[R|t]$ was derived by taking the image of the checkerboard and using its known position with respect to frame $\mathcal{A}_5/\mathcal{B}_5$ when it is rigidly mounted on the stator, as shown in Fig.~7(B).
We applied OpenCV \textit{calibrateCamera}~\cite{bradski2000opencv} to the image pixels and their corresponding 3D positions to get the intrinsic matrix $K$ and the extrinsic matrix $[R|t]$.

\textbf{3D Reconstruction}
The 3D positions of the points on the convex sensing area can be projected from the Cartesian space to the 2D camera image space using the camera matrices. Because the geometry of the roller is known, this projection also allows us to trace the 3D position of a point given its 2D coordinate in the image. This mapping of the points on the sensing area between their 3D positions and 2D image pixels are saved for 3D reconstruction when an external object is in contact with the roller. 

As shown in Fig.~8, when an object is in contact with the roller, the elastomer on the roller is deformed, creating a shaded image that is recorded by the camera. After unwarping the image into a rectangular shape (with the same pixel density along its horizontal and vertical axes), we applied photometric stereo to create a depth image: each pixel on the depth image will have a corresponding depth value, indicating the offset from its position on the undeformed roller surface. We apply this depth image on top of the mapping described previously to reconstruct the 3D geometry of the contacted object. This is accomplished by subtracting the offset of each pixel in the depth image from its corresponding 3D position along the surface normal direction. 

The photometric stereo technique used in this work is developed based on previous work in 3D reconstruction using a planar elastomer. Specifically, we first transformed the shaded image into surface normals, and then applied the fast Poisson solver~\cite{fastPoisson} for integration to produce the depth image. Further details of this method can be found in~\cite{yuan2017gelsight}.

\textbf{Marker Tracking}
The shear force estimation can be obtained by motion analysis, i.e., analyzing the marker displacement on the sensor in comparison with the reference images. During operation, markers are constantly disappearing and appearing from the boundaries of the sensor image due to the rotation of the roller, making the calculation of the marker displacement field difficult. A sensor image might even have a different number of markers compared to its reference image because certain markers are located at the very edge of the sensing area. Such a problem becomes especially prominent when the sensing area is constantly moving during the rolling motion. In these situations, techniques using marker tracking with nearest temporal matching~\cite{yuan2017gelsight,yamaguchi2016combining} or optical flow~\cite{sferrazza2019design,zhang2022deltact} tend to generate erroneous results.
Instead, we adopted Random Optimization to reliably track marker displacement during rolling by maximizing for marker flow smoothness~\cite{horn1981determining}, which assumes that nearby markers move with similar velocities, as well as minimizing marker mismatch between frames.
%
%
%

More formally, we define marker locations in the reference frame as $\{\bm{u}'_1, \bm{u}'_2, ..., \bm{u}'_{s}\}$, and marker locations in the current frame as $\{\bm{u}_1, \bm{u}_2, ..., \bm{u}_k\}$, where $\bm{u}',\bm{u}\in \mathbb{R}^2$ represent the markers' pixel coordinates, and $s$ is not necessarily equal to $k$. 
The goal of marker tracking is to find the correspondence between $\{\bm{u}'_i\}$ and $\{\bm{u}_j\}$. 
For each $\{\bm{u}'_i\}$,
we define $m_i$ as the index of its corresponding marker in the current frame, i.e., the marker at $\bm{u}_{m_i}$ overlaps with $\bm{u}'_i$ when the roller is untouched, 
and $m_i = -1$ when it cannot find a match in the current frame.

The commonly used nearest matching algorithm \cite{yuan2017gelsight,yamaguchi2016combining}, 
tends to converge at locally optimal solutions,
%
%
which causes mismatches for large marker displacements or markers that move out of frame.
Instead, we used an optimal matching algorithm that prioritizes the flow smoothness despite the high possibility of marker mismatches introduced by the rolling mechanism.

For simplicity, we define the indices for the neighboring markers of $\bm{u}'_i$ as the set $N_i=\{n_{ij}\}$, where each neighboring marker at $\bm{u}_{n_{ij}}$ is within a certain distance from $\bm{u}'_i$. Most $\bm{u}'_i$ have four neighboring markers, but those near the image boundary or those that are cluttered can have different numbers of neighbors.
As defined in the following equation, $d_i$ is the displacement between $\bm{u}'_i$ and $\bm{u}_{m_i}$, and the displacement of an unmatched marker is calculated by averaging the displacements of its neighboring markers. 

\begin{equation}
    \bm{d}_i = 
    \begin{cases}
        \bm{u}_{m_i} - \bm{u}'_i,& \text{if } m_i > 0\\
        \frac{1}{\vert N_i \vert}\sum_{n_{ij} \in N_i} \bm{d}_{n_{ij}} \cdot \mathbbm{1}\{m_{n_{ij}}>0\},              & \text{if } m_i = -1
    \end{cases}
\end{equation}

where $\mathbbm{1}$ is the indicator function.

We formulate the task as an optimization problem, with the goal to minimize the loss $L$, which consists of the smoothness term $L_{smooth}$ and the mismatching term $L_{mismatch}$, where $L_{smooth}$ considers the differences between neighboring marker displacements, inspired by the smoothness objective in optical flow~\cite{horn1981determining}, and $L_{mismatch}$ provides penalties for marker mismatch:

\begin{align*}
    L &= L_{smooth} + L_{mismatch}   \\
    L_{smooth} &= \sum_{i=1}^{s} \sum_{n_{ij} \in N_i} \lVert \bm{d}_i - \bm{d}_{n_{ij}} \rVert\\
    L_{mismatch} &= \sum_{i=1}^{s} K_{mismatch} \cdot \mathbbm{1}\{m_i = -1\}
\end{align*}

where $K_{mismatch}$ is a hyper-parameter to determine the significance of the mismatch objective compared to the smoothness objective.

Minimizing $L$ is a combinatorial optimization problem and is computationally expensive to exhaust all solutions. 
Therefore, we apply Random Optimization to heuristically generate several solutions in a limited period, and choose the one with minimal loss.
We sequentially assign the corresponding marker for each $\bm{u}'_i$ in a stochastic mean. First, we calculate the matching probability $p_{ij}$ for $\bm{u}'_i$ and $\bm{u}_j$, which is subsequently normalized and used to sample $m_i$. To ensure each marker is matched at most once, we set $p_{ij}=0$ if $\bm{u}_j$ is previously matched. The following heuristic is defined to ensure a higher matching probability for closer corresponding markers:

\begin{equation}
    p_{ij} = 
    \begin{cases}
        1 - \sigma(\lVert \bm{u}'_i - \bm{u}_j \rVert - T),& \text{if } j > 0\\
        1 - \sigma(T),  & \text{if } j = -1
    \end{cases}
\end{equation}

where $\sigma$ is the sigmoid function to map the marker distance to 0-1, and $T$ is a hyperparameter to determine the marker displacement range.

Based on this formulation, prior to normalization, the matching probability $p_{ij} > 0.5$ when $\lVert \bm{u}'_i - \bm{u}_j \rVert < T$, and $p_{ij} < 0.5$ otherwise. The condition $j = -1$ represents the situation where no marker is matched. This happens when all $\bm{u}_j$ have a distance larger than $T$ from $\bm{u}'_i$, indicating the corresponding marker is not detected in the current frame.
This heuristic random matching is repeated multiple times to find the minimal loss $L$. 

%
%

Unlike the exhaustive search, this method is suitable for real-time signal processing. 
With 200 possible solutions sampled for each frame, the algorithm can achieve real-time marker tracking at the frequency of $30$ Hz.

\subsection*{Control methods for In-Hand Manipulation}

We developed a series of manipulation tasks for the TRRG to demonstrate its capabilities. While these demonstrations required various high-level control methods, the low-level joint space control method is consistent across all of them. The base joints used current-limited position control to ensure that the object is being grasped securely without generating excessive internal forces. Position control is used to drive the pivot angle between $\pm 90^{\circ}$. Smooth rolling motion is achieved through velocity control of the rollers. 
For each manipulation task presented in this work, we used the tactile sensor to close the control loop. Fig.~6(B) shows the control diagram. The high-level algorithm can be any heuristic or autonomous planner that outputs the desired object velocity $v_{obj,d}$, which is mapped to the desired pivot angles $\theta_{p,d}$ and roller velocities $\omega_{r,d}$ in the low-level controller. The sensed contact location was sent back to the low-level controller while the high-level controller receives both the contact location and estimated shear force. The surface coordinate information extracted from the sensor measurement was fed through the modified contact equations derived from~\cite{montana1988kinematics}.

The control loop is closed by inferring the object velocity based on the contact coordinates through differential geometry. 
Consider the shape of the roller shown in Fig.~3(G). We can define the coordinate patch $f$ that maps the open set $U \subset \mathbb{R}^2$ to the roller surface $S \subset \mathbb{R}^3$, i.e., $f$ is a map for $U\to S \subset \mathbb{R}^3$ for all the points on the surface with coordinate $\bm{u}=(u,v)\in U$. The surface $S$ is defined with respect to frames $\mathcal{A}_6/\mathcal{B}_6$. Based on the geometry of the roller and considering the set $U=\{(u,v) | -5\pi/12 < u < -5\pi/12, -\pi < v< \pi\}$,  the map can be defined as:
\begin{equation}
    (u,v) \mapsto (  (R_0\cos u-D_0)\cos v, -(R_0\cos u - D_0)\sin v, R_0\sin u   )
\end{equation}
Note that this formulation can be used for any roller form-factor in the previous Roller Graspers \cite{rgv1,rgv2}. When $D_0=0$, the formulation becomes the mapping for spherical rollers, while as $u\to 0$, the formulation becomes the mapping for cylindrical rollers. 
Based on (7)-(9) in~\cite{montana1988kinematics}, the curvature form $K$, the torsion form $T$, and the metric $M$ can be derived accordingly as
\begin{equation}
    K = \begin{bmatrix}
        \frac{1}{R_0} & 0 \\
        0 & \frac{\cos u}{R_0\cos u - D_0}
    \end{bmatrix} 
    \text{ and  }
    T = \begin{bmatrix}
        0 & \frac{\sin u}{D_0 -R_0 \cos u}
    \end{bmatrix} 
    \text{ and  }
    M = \begin{bmatrix}
        R_0 & 0 \\
        0 & R_0\cos u - D_0
    \end{bmatrix} 
\end{equation}
Assuming a known object geometry around the contact location, a point contact between the object and the roller, and rolling without slipping, the rotational velocity of the object can be calculated as 
\begin{equation}
\begin{bmatrix}
    -\omega_y   \\ \omega_x
\end{bmatrix}
=(K+\widetilde{K}_{obj} )M \dot{\bm{u}} = (K+\widetilde{K}_{obj} )M (\dot{\bm{u}}_s - \dot{\bm{u}}_r)
\end{equation}
We define the contact frame on the roller as right-handed with its origin at the contact point, the $z$-axis normal to the surface pointing outward, and the $x$-axis and $y$-axis pointing in the longitudinal and latitudinal directions of the roller, respectively.  $\omega_x$ and $\omega_y$ are the components of the object's rotational velocity with respect on the contact frame on the roller. $\widetilde{K}_{obj} = R_{\psi} K_{obj} R_{\psi} $ is the curvature of the grasped object relative to the contact frame on the roller, where $R_{\psi}$ defines the rotation between the contact frame on the roller and that on the object, and $K_{obj}$ is the curvature form of the object.  $\dot{\bm{u}}_r = [\omega_r, 0]$ is the difference between $\dot{\bm{u}}_r$ and $\dot{\bm{u}}_s$ caused by the roller speed, where $\omega_r$ is the real-time roller speed.



Next, we describe specific manipulation tasks to demonstrate the capabilities of the TRRG.  

\textbf{Cylindrical object rotation} As shown in Fig.~4(A), the controller adjusts the rolling speed of the fingertips to keep the pen rotating at a specified speed. We found the controller is robust even in the presence of external disturbances.
In addition, we can extract the primary and secondary principal axes of the contact area using principal coordinate analysis~\cite{she2021cable}, which, for an object with a relatively consistent contact shape indicate the orientation of the contact geometry, and subsequently the pose of the grasped object. In this particular example, the principal axis indicates the long axis of the pen.

\textbf{Planar object reorientation} The planar object reorientation (Fig.~4(B)) attempts to keep the object spinning around a horizontal axis. 
The planar objects used in this demonstration are 3D printed with varying radii of curvature to demonstrate that the control method can adapt to complex and unknown 2D geometries. Tactile sensing was used to adjust the pivot angle of the rollers to stably hold the object during the manipulation.
Each of the previous two examples were also run in open-loop without the sensor feedback, and the grasped objects were dropped shortly after the experiment began. 

\textbf{Spherical object screw motion and trajectory following} Unlike the previous two demonstrations with a fixed control target, this task (Fig.~4(C)) attempts to move the object along pre-defined trajectories in the operational space. Object position is computed using forward kinematics, and used to close the high-level trajectory following control loop. 
The screw motions achieved in this demonstration, where the translational and rotation motions are coupled, are difficult or impossible to perform with traditional robot hands. The TRRG easily achieves screw motions by setting the rollers in opposite orientations, forming a cross. Changing angle between the rollers enables setting the screw pitch from zero to infinity, and anywhere in between.


\textbf{Cable tracing} Unlike rigid body objects, cables can withstand substantial tensile load but can buckle under modest compressive loads. This makes a cable a very difficult object for robot manipulation. 
Tracing along the length of a cable can only be achieved while sliding away from where the cable's anchor in order to maintain the cable tension as was shown in ~\cite{she2019cable}; moving towards the anchored location which will cause the cable to buckle due to friction. In our case, the combination of the rolling motion and the ability to control the estimated shear force applied to the cable makes it possible for the rollers to trace along a cable both toward and away from the anchor point while maintaining the cable tension. The contact location can also be computed to prevent the cable from dropping due to gravitational forces; by adjusting the pivot direction the contact location can be kept in the center of the roller. We tested two open-loop cases for this demonstration. The first case does not use any sensor feedback, which results in the cable being dropped almost immediately due to gravity.
The second case tracks the contact location but ignores the shear information. The rollers are able to move along the cable for longer, however, the cable eventually became slack and was no longer traversable. Using both the contact location and shear force information, the cable could be traversed without fail. 

\textbf{Card picking} Another way of using the tactile information is to capture the transient events, something done by humans in various manipulation tasks. In particular, transient events can be useful when interfacing with thin objects, like playing cards. Their extreme aspect ratios increase the likelihood for several cards to stick together, therefore, distinguishing between multi-card and single-card grasps can be extremely difficult. The TRRG can circumvent this by using a single roller to actuate one side of a given card and monitor the observed shear force. 
When multiple cards are within the hand, relative motion of any two cards reduces the experienced shear force on the roller. However, sudden changes in the shear force indicate that only one card is left in grasp. This is a specific example demonstrating the use of transient shear information, but in practice, there are various situations where this methodology can be used~\cite{veiga2015stabilizing,chen2018tactile,kuppuswamy2020soft}, especially when the state of the hand-object configuration experiences sudden changes.



\FloatBarrier

\section*{Supplementary Materials and Methods}

Texts S1 to S2
Figs. S1 to S2
Table S1
Movies S1 to S2

\FloatBarrier

\bibliographystyle{IEEEtran}
\bibliography{scibib}

\FloatBarrier

\section*{Acknowledgments}
The authors would like to thank Sandra Liu for setting up the laser cutter with the rotary attachment. \textbf{Funding:} This work was supported in part by Toyota Research Institute (TRI), the Office of Naval
Research (ONR) [N00014-18-1-2815], and GentleMAN and BiFrost projects of SINTEF. Shenli Yuan was supported by the Stanford Interdisciplinary Graduate Fellowship.  \textbf{Author contributions:} Shenli Yuan designed hand mechanism, participated in the sensor design, designed the electronics and firmware, developed low-level control software as well as part of manipulation software, and built a hand prototype. Shaoxiong Wang led the sensor design, performed signal processing for sensor images, developed part of the manipulation software and the object reconstruction software, and built a hand prototype. Radhen Patel fabricated roller elastomer, and participated in prototype assembly. Megha Tippur fabricated roller elastomer and designed encoder markers. Connor Yako participated in mechanism design and prototype assembly. Edward Adelson and Kenneth Salisbury are advisors for the project. \textbf{Competing interests:} The authors declare that have no competing interests. \textbf{Data and materials availability:} All data needed for the evaluation of the paper are present in the paper and the supplementary materials. 

\FloatBarrier



\begin{figure}
    \centering
    \includegraphics[width=7in]{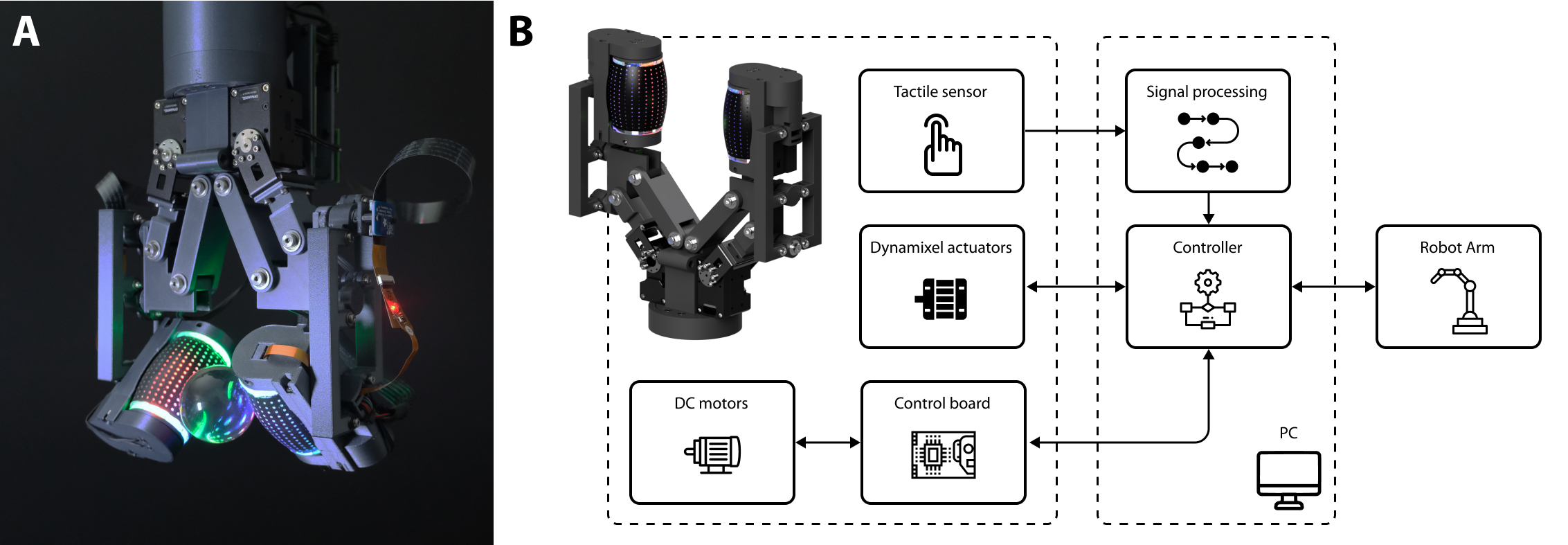}
    \caption{
    \textbf{TRRG prototype and system.}
(\textbf{A}) A physical prototype of the TRRG manipulating a clear ball. 
(\textbf{B}) System setup consisting of the TRRG, a PC and a UR5e robot arm. The PC handles the tactile signal processing and controls the TRRG and the robot arm. 
    }
    \label{fig:system}
\end{figure}

\begin{figure}
    \centering
    \includegraphics[width=6in]{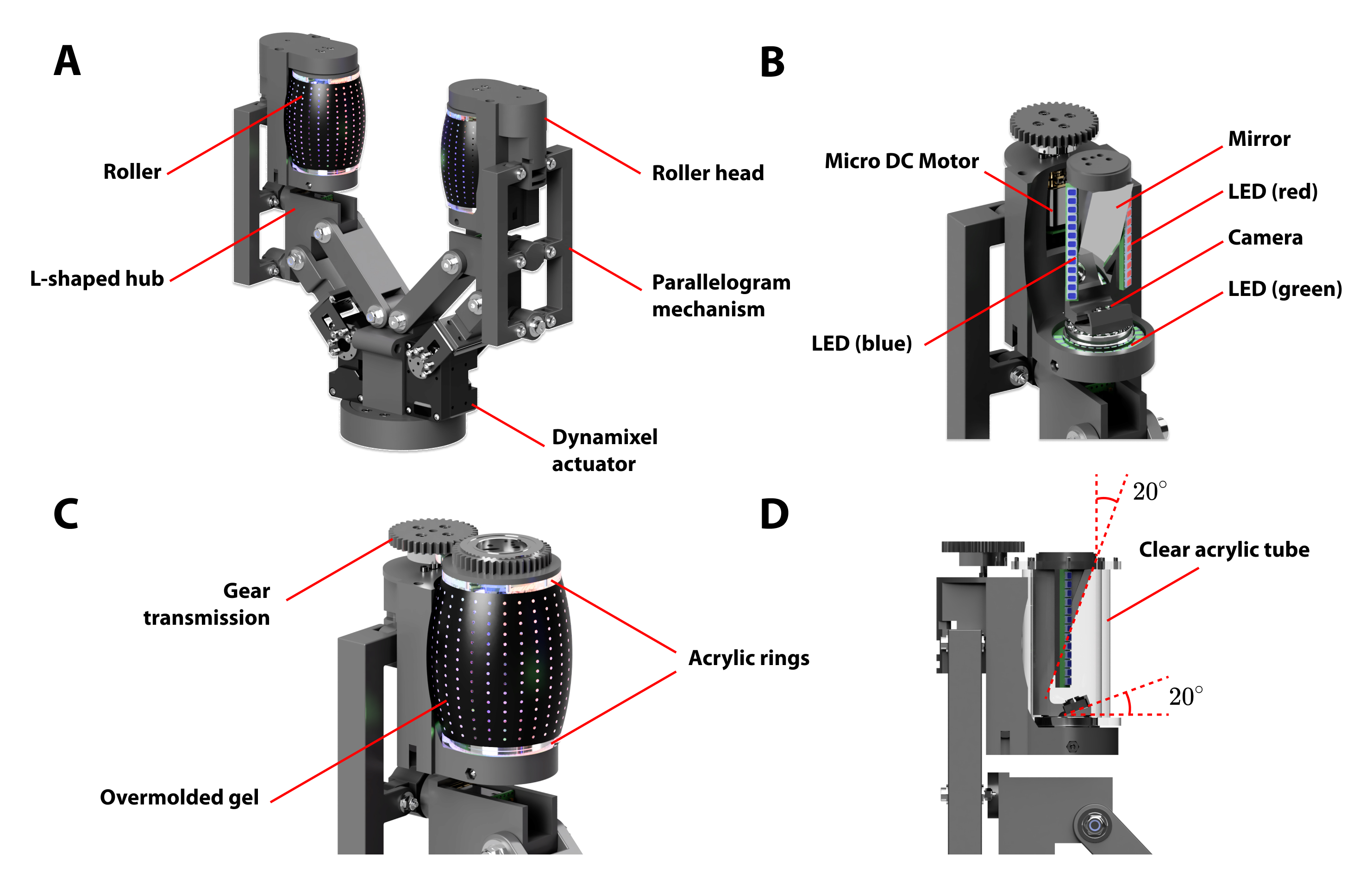}
    \caption{
    \textbf{CAD renderings of the mechanical design of the TRRG.} (\textbf{A}) Fully assembled hand. (\textbf{B}) Non-rotating optical components (stator) inside the roller, the rotating part is called the rotor. (\textbf{C}) The roller assembly and drive motor. (\textbf{D}) The camera and mirror in the roller.
    }
    \label{fig:render}
\end{figure}

\begin{figure}
    \centering
    \includegraphics[width=6in]{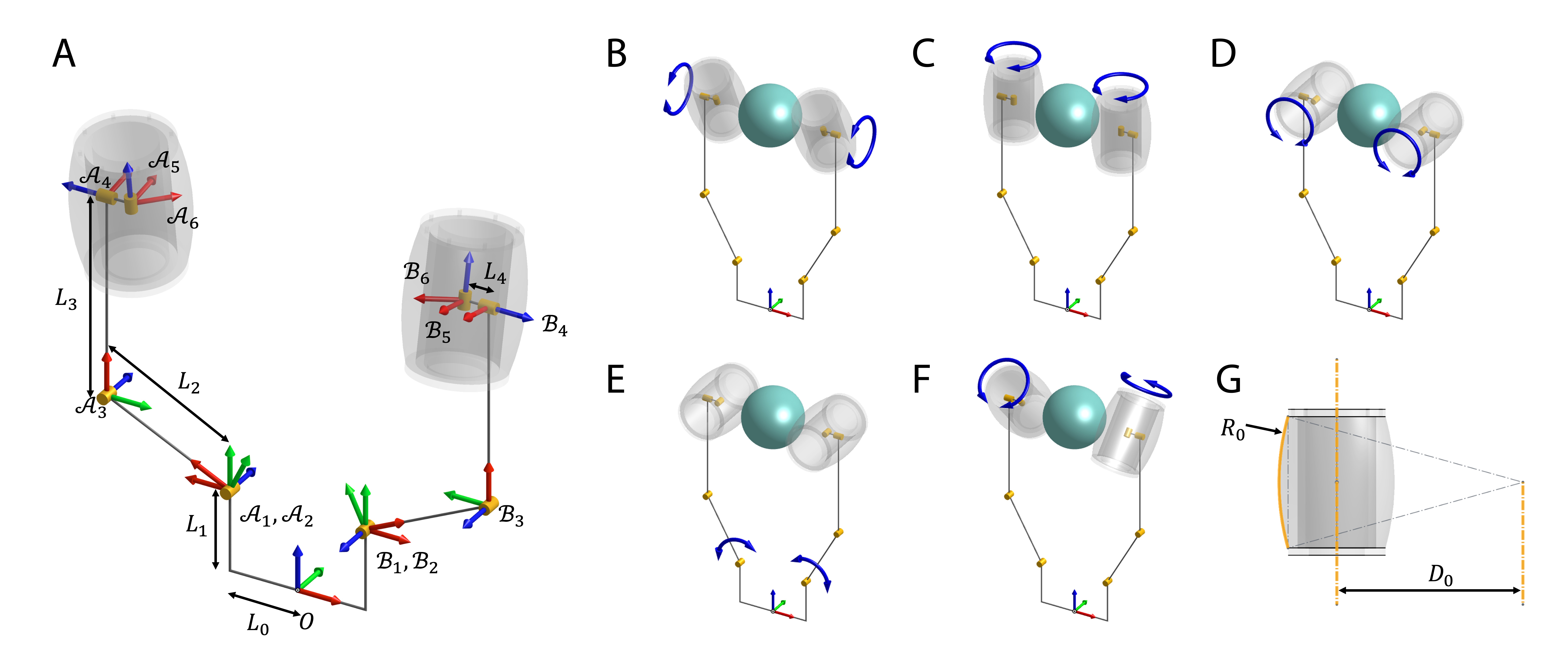}
    \caption{\textbf{3D kinematics and roller configurations.} (\textbf{A}) The TRRG frame definitions. $\mathcal{A}$ and $\mathcal{B}$ represent each finger. Frame $O$ is the fixed frame located at the base of the hand. The numerical subscripts represent frames attached at different locations of the hand. 
The $X, Y, Z$ axes are represented in red, green and blue, respectively. Frame $O$ is the world frame with which we reference the manipulation directions. (\textbf{B}) Object rotation in $X_O$. (\textbf{C}) Object rotation in $Z_O$ or object translation in $Y_O$, depending on the rolling directions of the two rollers. (\textbf{D}) Object rotation in $Y_O$ or object translation in $Z_O$, depending on the rolling directions of the two rollers. (Any rotation or translation in directions within $Y_O-Z_O$ plane are possible with different pivot positions) (\textbf{E}) Object translation in $X_O$. (\textbf{F}) Object screw motion (coupled rotation and translation. (\textbf{G}) Roller geometry  
    }
    \label{fig:mobility}
\end{figure}

\begin{figure}
    \centering
    \includegraphics[width=5.in]{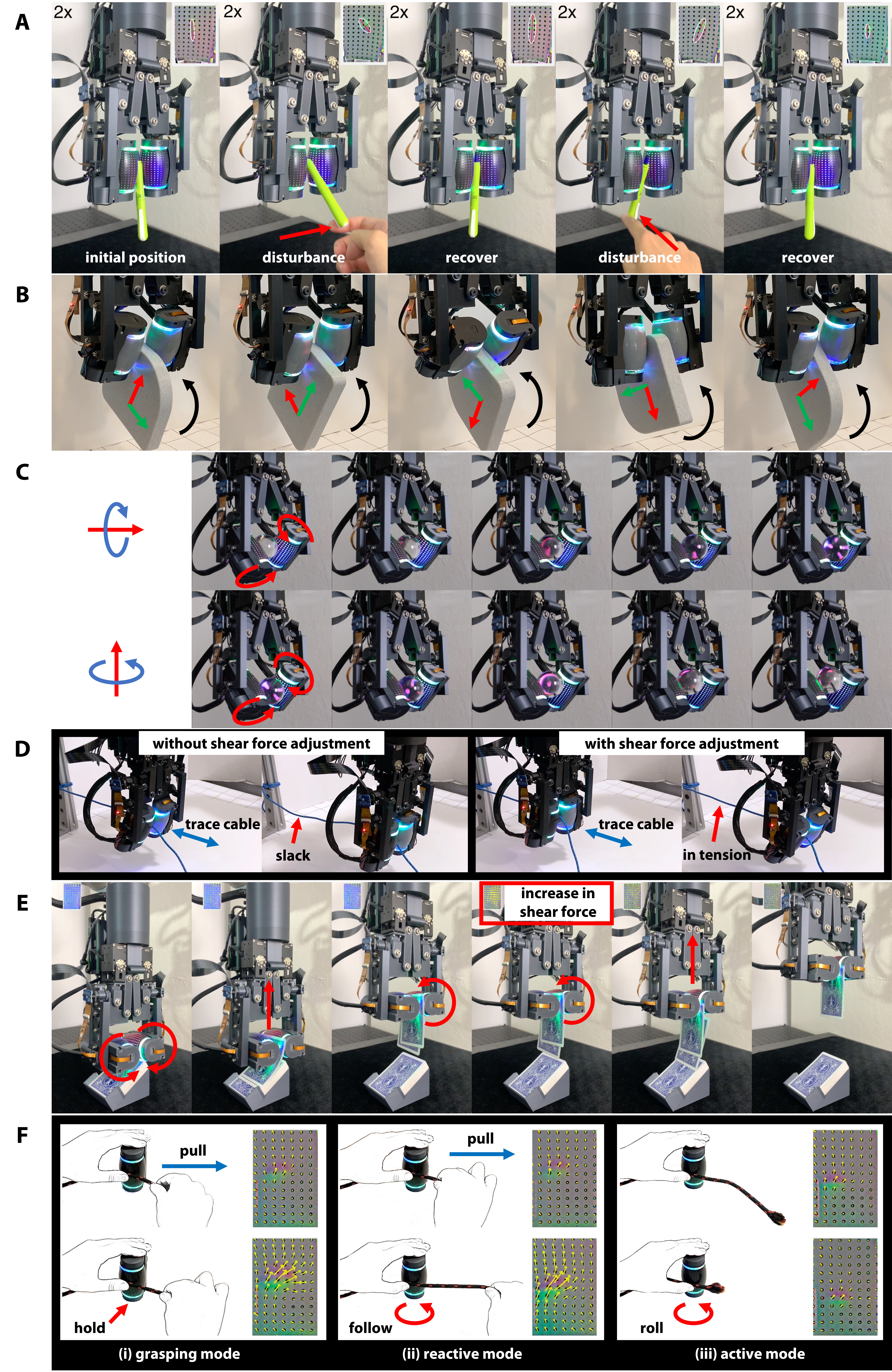}
    \label{fig:demo_combined}
\end{figure}
\addtocounter{figure}{0}
\begin{figure} [t!]
  \caption{
    \textbf{In-hand manipulation demonstrations.}
    (\textbf{A}) Cylindrical object manipulation. 
    (\textbf{B}) Planar object manipulation. 
    (\textbf{C}) Spherical object manipulation. 
    (\textbf{D}) Cable manipulation: The roller grasper can stably traverse bidirectionally along a cable. 
    Closed-loop control for the sensed shear force and the contact location maintain the cable tension (Right) and compensate for disturbances from gravitational forces. Without controlling the shear force the cable accumulates slack (Left).
    (\textbf{E}) Card picking: The TRRG picks up a single card by actuating a single roller and monitoring transient shear force signals. These signals help distinguish between multi-card and single-card grasps.
    (\textbf{F}) Different roller actuation modes. (\textbf{i}) Grasping mode: the roller is non-backdrivable, and will resist external force due to friction. (\textbf{ii}) Reactive mode: software backdriving. The roller can react according to the amount of shear force applied on the roller surface (\textbf{iii}) Active mode: the roller actively apply motions to the contacted object. 
    }
\end{figure}

\begin{figure}
    \centering
    \includegraphics[width=6in]{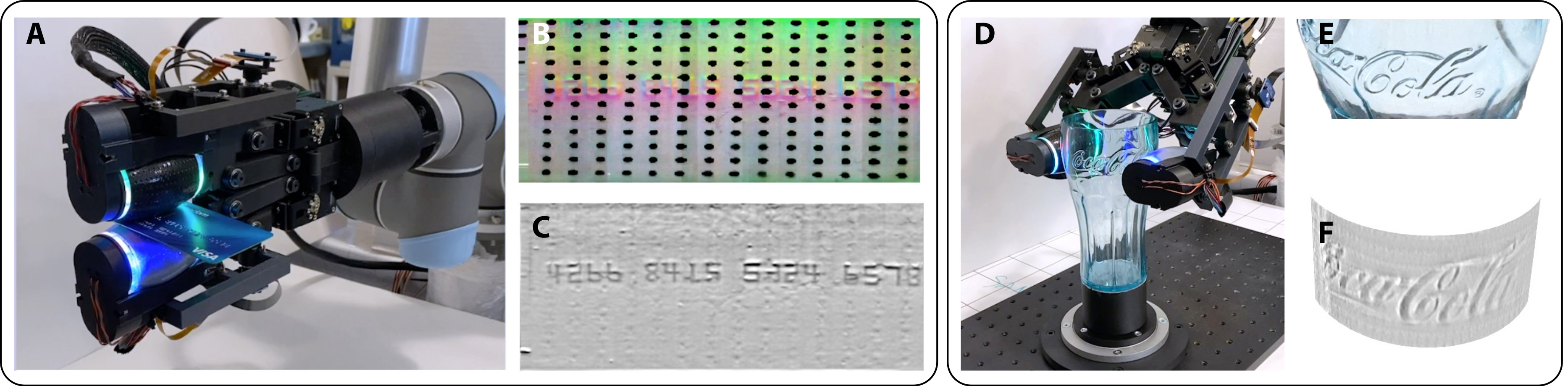}
    \caption{
    \textbf{Surface scanning.} 
   (\textbf{A}) Rolling along a credit card. 
   (\textbf{B}) Stacked tactile images in the time sequence, showing the embossed numbers on the credit card. 
   (\textbf{C}) Processed tactile image with interpolation at the marker region and sharpening filters for better visualization. 
   (\textbf{D}) Rolling along a transparent cup. 
   (\textbf{E}) The embossed characters on the cup. 
   (\textbf{F}) Scanned tactile images stitched in 3D spaces.
   }
    \label{fig:inspection}
\end{figure}

\begin{figure}
    \centering
    \includegraphics[width=6in]{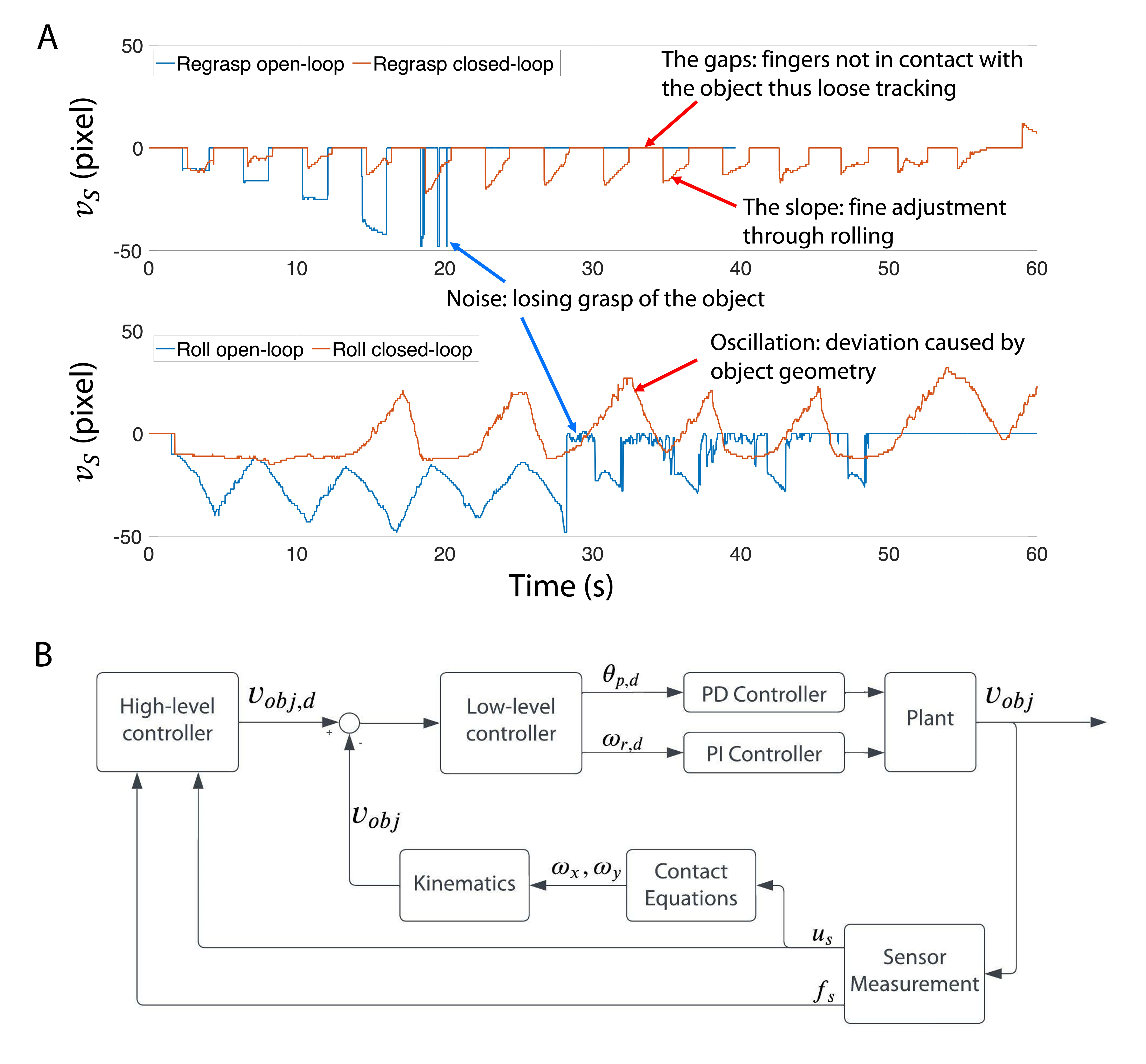}
    \caption{ \textbf{Experiment and control diagram.}(\textbf{A}) Results of the experiment comparing rolling vs. regrasping. The data represent the horizontal coordinate of the contact location recorded by the tactile sensor. 
    (\textbf{B}) Control Diagram. The heuristic-based high-level controller outputs the desired object velocity, which is mapped to the joint velocities in the low-level controller. Both controllers receive sensor measurements as feedback.   }
    \label{fig:system}
\end{figure}

\begin{figure}
    \centering
    \includegraphics[width=5in]{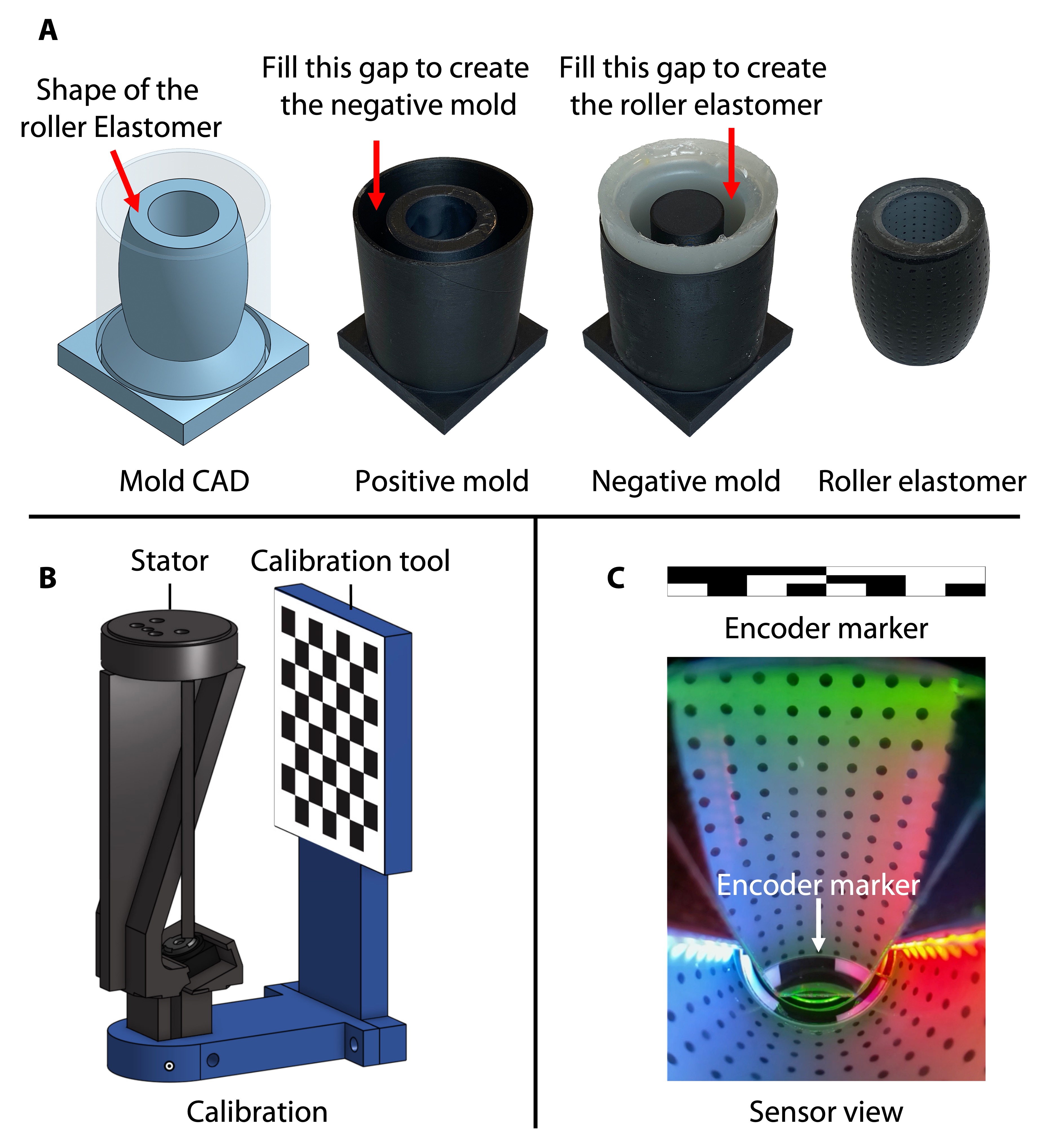}
    \caption{(\textbf{Sensor Construction and Calibration.}(\textbf{A}) Mold for the seamless roller elastomer. From \textit{Left} to \textit{Right}: CAD model of the mold; 3D printed positive mold, with surface smoothed; rubber negative mold; seamless elastomer covering around clear acrylic tube. 
    (\textbf{B}) A 7x8 Checkerboard mounted on a calibration tool to get the sensor intrinsic and extrinsic matrices. 
    (\textbf{C}) The pattern of the encoder marker to provide precise position encoding, and the corresponding image from the sensor view.}
    \label{fig:mold}
\end{figure}

\begin{figure}
    \centering
    \includegraphics[width=6in]{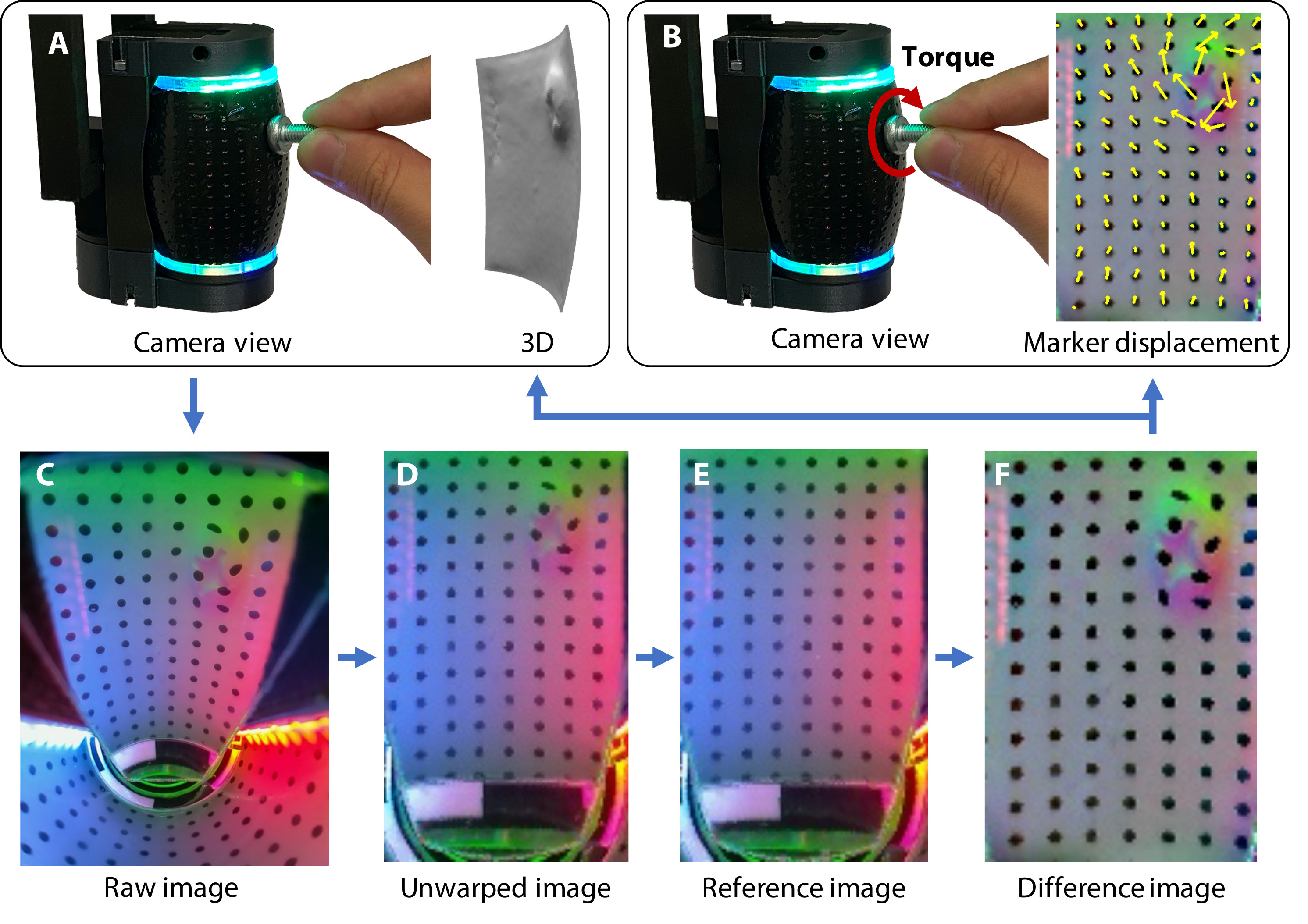}
    \caption{\textbf{3D reconstruction and marker tracking.} (\textbf{A}) The camera view shows a screw head pressing on the roller sensor, and the 3D view shows the estimated 3D reconstruction. (\textbf{B}) The camera view demonstrates the torque exerted on the roller sensor, and the marker displacement visualizes the magnified motion of the markers captured from the sensor.  (\textbf{C}) The camera inside the roller sensor captures the raw image, and the sensing area is captured in the mirror. (\textbf{D}) The raw image is unwarped into a rectangular image. (\textbf{E}) The reference image is extracted with the encoder marker from the unwarped image. (\textbf{F}) The difference image is calculated between the unwarped image (after contact) and the reference image (before contact). It is further processed to get the 3D reconstruction and marker displacement.}
    \label{fig:Perception}
\end{figure}

\FloatBarrier

\begin{table}
\centering
\caption {\textbf{Physical properties of the TRRG}: Labels shown in Fig. 3} 
\begin{tabular}{|c|c|c|c|c|c|c|}

\hline
Properties  & $L_0$ (mm)  &  $L_1$ (mm) & $L_2$ (mm) & $L_3$ (mm)     & $R$ (mm)  & $D$ (mm)  \\ \hline
Value       & $29.75$     & $35.25$     & $60$         & $87.25$       & $100$        & $80$   \\ \hline

\end{tabular}
\label{table:control_summary}
\end{table}




\end{document}